\documentclass[letterpaper,journal]{IEEEtran}
\IEEEoverridecommandlockouts
\usepackage{amsmath,amssymb,amsfonts}
\usepackage{algpseudocode}
\usepackage{algorithm}
\usepackage{graphicx}
\usepackage{textcomp}
\usepackage{xcolor}
\usepackage{orcidlink}
\usepackage[caption=false,font=normalsize,labelfont=sf,textfont=sf]{subfig}
\usepackage[inline]{enumitem}
\usepackage[style=ieee, doi=false, url=false, date=year, isbn=false]{biblatex}
\AtNextBibliography{\small}
\AtEveryBibitem{\clearfield{pages}}
\AtEveryBibitem{\clearfield{pagetotal}}
\AtEveryBibitem{\clearfield{volume}}
\AtEveryBibitem{\clearfield{number}}
\AtEveryBibitem{\clearlist{language}}
\def\BibTeX{{\rm B\kern-.05em{\sc i\kern-.025em b}\kern-.08em
    T\kern-.1667em\lower.7ex\hbox{E}\kern-.125emX}}
\begin{document}

\title{Simultaneous Genetic Evolution of Neural Networks for Optimal SFC Embedding}

\author{
Theviyanthan Krishnamohan\textsuperscript{\orcidlink{0000-0003-0040-1130}}, Lauritz Thamsen\textsuperscript{\orcidlink{0000-0003-3755-1503}}, and
Paul Harvey\textsuperscript{\orcidlink{0000-0003-1243-938X}}
}

\maketitle

\begin{abstract}
The reliance of organisations on computer networks is enabled by network programmability, which is typically achieved through Service Function Chaining. These chains virtualise network functions, link them, and programmatically embed them on networking infrastructure. Optimal embedding of Service Function Chains is an $\mathcal{NP}$-hard problem, with three sub-problems---chain composition, virtual network function embedding, and link embedding---that have to be optimised simultaneously, rather than sequentially, for optimal results. Genetic Algorithms have been employed for this, but existing approaches either do not optimise all three sub-problems or do not optimise all three sub-problems simultaneously.

We propose a Genetic Algorithm-based approach called GENESIS, which evolves three sine-function-activated Neural Networks, and funnels their output to a Gaussian distribution and an A* algorithm to optimise all three sub-problems simultaneously. We evaluate GENESIS on an emulator across 48 different data centre scenarios and compare its performance to two state-of-the-art Genetic Algorithms and one greedy algorithm. GENESIS produces an optimal solution for 100\% of the scenarios, whereas the second-best method optimises only 71\% of the scenarios. Moreover, GENESIS is the fastest among all Genetic Algorithms, averaging 15.84 minutes, compared to an average of 38.62 minutes for the second-best Genetic Algorithm. 
\end{abstract}


\section{Introduction}
\IEEEPARstart{c}{omputer} networks underpin numerous facets of modern life, from facilitating critical services like healthcare~\cite{Pandav2022LeveragingCare.} and finance~\cite{annurev:/content/journals/10.1146/annurev-financial-101620-063859} to enabling recreational activities such as gaming~\cite{6727567} and video streaming~\cite{10.1145/3519552}.
Accordingly, the safe, reliable, and efficient operation of computer networks is essential for maintaining a functioning society. This requires computer networks to keep pace with the changing ways in which they are relied upon~\cite{Imai2020TowardsNetwork}. 

Network programmability is the most promising approach for keeping up with the pace of change~\cite{Imai2020TowardsNetwork}. Traditional networks require manual human intervention to adapt to changes in the operating environment, which hinders faster adaptation. By enabling network programmability, computer networks can be programmatically configured and manipulated, minimising manual intervention and enabling faster adaptation to changes.

Service Function Chaining (SFC)~\cite{GilHerrera2016} enables network programmability by combining Network Function Virtualisation (NFV)~\cite{NFV2014} and Software Defined Networking (SDN)~\cite{2013Software-DefinedIt}. NFV first separates network functions, such as a firewall, from hardware implementations by virtualising them, enabling them to be embedded programmatically on any physical host. SDN enables programmatic configuration of how traffic is routed between Virtual Network Functions (VNFs) by embedding virtual links between VNFs on physical links, creating a chain of VNFs. 

However, this programmability introduces more complexity and configuration options. Consider a network service provider in a data centre environment that receives several SFC Requests (SFCRs). SFCRs are requests to embed SFCs on a physical network, and the provider must embed them on their network, optimising objectives according to business needs, such as traffic latency and energy consumption. Each SFCR contains the VNFs in the SFCs and the order between some or all of the VNFs. The network provider requires an optimisation approach to find an optimal embedding scheme to embed the SFCRs they receive. We call this the Optimal Service Function Chain Embedding (OSE) problem~\cite{OpenRASE}. The OSE problem involves optimally ordering the VNFs in an SFC, called Chain Composition (CC), embedding the VNFs on hosts, called VNF Embedding (VE), and embedding the virtual links between VNFs on the physical network, called Link Embedding (LE). 

All three sub-problems have to be optimised simultaneously, rather than sequentially, to produce an optimal solution, which is an $\mathcal{NP}$-hard optimisation problem~\cite{GilHerrera2016}. Optimising all three sub-problems creates a solution space with more degrees of freedom, by configuring the order of VNFs, their hosts and the links between them, and optimising them simultaneously allows the entire search space to be explored. 

Genetic Algorithms (GAs)~\cite{558db5e0-7a5a-3da1-8b88-2f13b93dbaf1} are meta-heuristic algorithms that are effective at optimising $\mathcal{NP}$-hard problems. Of the 17 studies~\cite{Khoshkholghi2019, Carpio2017, 10.1007/978-3-031-33743-7_39, Toumi2022, Gamal2019, Gamal2019a, Qu2016, Cao2017, Rankothge2017, Khoshkholghi2020, Ruiz2020, Kim2016, Tavakoli-Someh2019, Fulber-Garcia2024BreakingOrchestrators, Fulber-Garcia2023CustomizableMetaheuristics, OpenRASE, Krishnamohan2025BeNNS:Embedding} in the literature using GAs to optimise at least one of the OSE sub-problems, only one~\cite{Cao2017} considered all three sub-problems, but it did so sequentially. Outside GAs, 11 studies optimised all three sub-problems simultaneously~\cite{Li2023a, 7417401, Sahhaf2015a, Jalalitabar2016ServiceDependence, Li2017ConstructingAllocation, Beck2017ScalableChains, Beck2016ResilientChains, Araujo2019AFunctions, Xu2018CoordinatedInfrastructure, Malektaji2023DynamicFramework, Sargolzaei2025TopologyNetworks}. However, only 6 of these studies proposed a scalable approach in the form of a heuristic or Reinforcement Learning approach~\cite{Li2023a, 7417401, Beck2017ScalableChains, Beck2016ResilientChains, Xu2018CoordinatedInfrastructure, Malektaji2023DynamicFramework}, but their LE approaches either ignore network switches~\cite{Li2023a, 7417401, Xu2018CoordinatedInfrastructure} or rely on a shortest-path algorithm~\cite{Beck2017ScalableChains, Beck2016ResilientChains, Malektaji2023DynamicFramework}, which could lead to congestion~\cite{Chetty2022}. 

In this paper, we propose a novel approach to OSE that simultaneously evolves three Neural Networks (NNs), using the sine activation function, named \textbf{G}enetic \textbf{E}volution of \textbf{Ne}ural Networks in \textbf{S}imultaneous \textbf{I}nteractive \textbf{S}olvers (GENESIS). We use the sine activation function to improve the exploratory behaviour of GENESIS. Each NN is contained within a solver, which uses the output of the NN to generate a candidate solution to an OSE sub-problem (Section \ref{GENESIS}). For instance, we use the output of one of the NNs to evolve an optimal Gaussian distribution, increasing exploration, to optimise VE (Section \ref{ve_solver}), while we use the output of another NN in the A* algorithm to find the optimal paths between hosts via switches (Section \ref{le_solver}). We use the hybrid online-offline evolution approach~\cite{Krishnamohan2025BeNNS:Embedding}, leveraging a surrogate model and an emulator to evolve the gradients of the NNs. To the best of our knowledge, this is the first paper that uses GAs to 
\begin{itemize}
    \item optimise all three OSE sub-problems simultaneously (Section \ref{sol_ose}),
    \item evaluate the fitness of such a solution on an emulator (Section \ref{sol_ose}),
    \item simultaneously evolve three NNs that use the sine activation function (Section \ref{ga_nn_rel}), and
    \item evolve a Gaussian distribution using the output of an NN (Section \ref{ga_nn_rel}).
\end{itemize}

We experimentally evaluate GENESIS across 48 different scenarios and compare its performance and scalability against two different GAs and a greedy algorithm from the literature 
(Section \ref{results}). GENESIS converged in 100\% of the scenarios, while the best among the other algorithms converged in only 71\% of the scenarios. GENESIS took an average of 15.84 minutes to converge across the 48 scenarios, while the fastest of the other GA-based algorithms averaged 38.62 minutes. 

\section{Related Work}
\subsection{Solutions to OSE}\label{sol_ose}
\subsubsection{Approaches to Optimise OSE}
The OSE problem has attracted significant research interest in recent years~\cite{Attaoui2023VNFTrends, Schardong2021, Kaur2020}. Our survey of the existing literature found that most research focused on optimising only one or two of the OSE sub-problems sequentially, with only 11 studies optimising all three OSE sub-problems simultaneously~\cite{Li2023a, 7417401, Sahhaf2015a, Jalalitabar2016ServiceDependence, Li2017ConstructingAllocation, Beck2017ScalableChains, Beck2016ResilientChains, Araujo2019AFunctions, Xu2018CoordinatedInfrastructure, Malektaji2023DynamicFramework, Sargolzaei2025TopologyNetworks}. Of these 11 studies, 5 proposed an exact solution~\cite{Sahhaf2015a, Jalalitabar2016ServiceDependence, Li2017ConstructingAllocation, Araujo2019AFunctions, Xu2018CoordinatedInfrastructure}. Exact solutions attempt to find the optimal solution by exploring all possible solutions. This works well when the search space is small, but does not scale well when the complexity of the problem increases~\cite{Toumi2022}. The remaining 6 studies employed a heuristic or Reinforcement Learning approach~\cite{Li2023a, 7417401, Beck2017ScalableChains, Beck2016ResilientChains, Xu2018CoordinatedInfrastructure, Malektaji2023DynamicFramework}, enabling their approach to scale well. 
\begin{figure}
    \centering
    \includegraphics[width=1\linewidth]{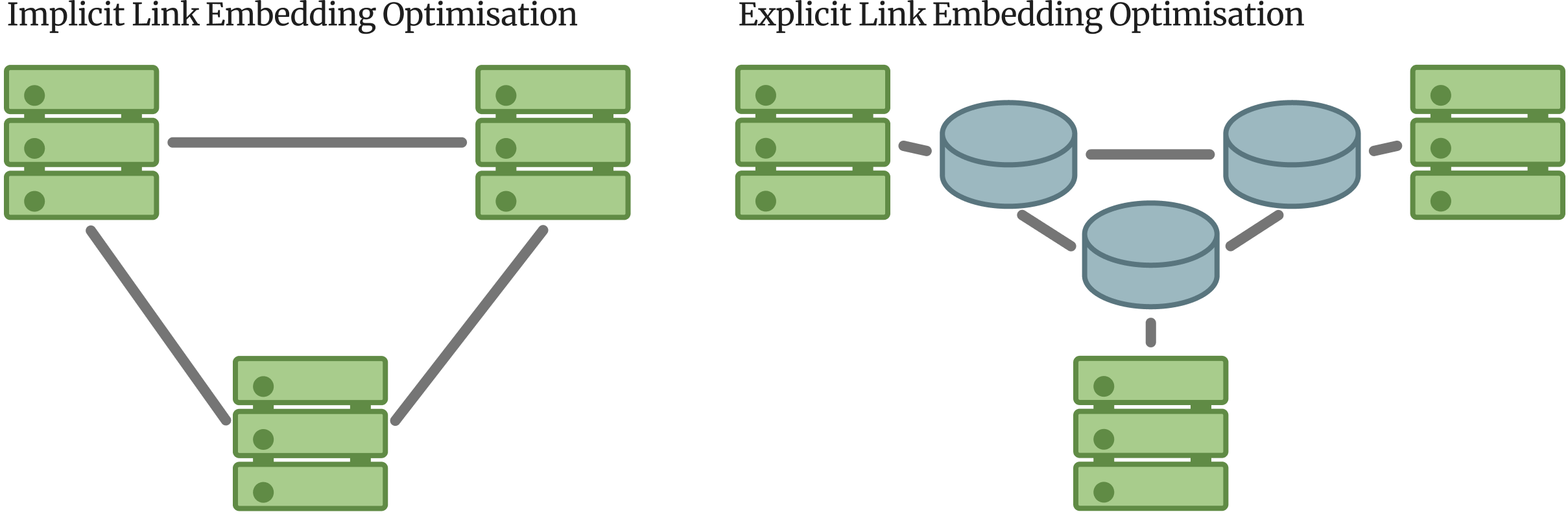}
    \caption{In implicit Link Embedding optimisations, studies assume that there are direct physical paths between hosts, ignoring the network switches. In explicit Link Embedding optimisations, studies consider the network switches and the possibility of multiple paths connecting hosts.}
    \label{fig:implicit_explicit}
\end{figure}

However, a recurring theme in the literature across all approaches is that they abstract away the complexity of the LE problem, which deals with finding optimal paths between hosts via network switches, by assuming that there exists a direct path between hosts~\cite{Li2023a, 7417401, Xu2018CoordinatedInfrastructure}. As shown in Fig. \ref{fig:implicit_explicit}, in network topologies, such as the fat tree topology~\cite{Leiserson1985Fat-Trees:Supercomputing}, multiple paths exist between hosts. Abstracting such complex topologies to direct paths between hosts ignores realistic traffic steering via switches. We consider such abstract optimisation approaches to LE to be \textit{implicit} optimisation. Other studies use different shortest-path algorithms, such as Dijkstra, to find the shortest path between hosts~\cite{Beck2017ScalableChains, Beck2016ResilientChains, Malektaji2023DynamicFramework}. Since such algorithms find the shortest path between hosts via switches among many paths, we consider such optimisation approaches to be \textit{explicit}. Of the 6 studies that took a scalable approach to the OSE problem, 3 studies~\cite{Li2023a, 7417401, Xu2018CoordinatedInfrastructure} used an implicit optimisation approach, whereas the other 3~\cite{Beck2017ScalableChains, Beck2016ResilientChains, Malektaji2023DynamicFramework} used an explicit approach. However, using shortest-path algorithms to link hosts ignores the traffic congestion caused by overloading links~\cite{Chetty2022}.

\subsubsection{GA-based Approaches to OSE}
We found 17 studies that used GAs to optimise the OSE problem~\cite{Khoshkholghi2019, Carpio2017, 10.1007/978-3-031-33743-7_39, Toumi2022, Gamal2019, Gamal2019a, Qu2016, Cao2017, Rankothge2017, Khoshkholghi2020, Ruiz2020, Kim2016, Tavakoli-Someh2019, Fulber-Garcia2024BreakingOrchestrators, Fulber-Garcia2023CustomizableMetaheuristics, OpenRASE, Krishnamohan2025BeNNS:Embedding}. Out of the 17 studies, only one optimised all three problems, but sequentially~\cite{Cao2017}.  Khoshkholghi et al.~\cite{Khoshkholghi2019} used GA to optimise the cost of an SFC embedding by optimising the VE problem. Link cost was minimised in the study by Carpio et al., which optimised VE, and LE implicitly~\cite{Carpio2017}. In addition to GA, the study also used Integer Linear Programming (ILP) and Random Fit Placement Algorithm for optimisation. Bekhit et al.~\cite{10.1007/978-3-031-33743-7_39} performed multi-objective optimisation by minimising bandwidth usage and delay cost, and maximising average resource usage across Virtual Machines. Toumi et al.~\cite{Toumi2022} simultaneously optimised VE and LE, taking an implicit approach, by minimising end-to-end latency and overall cost, while maximising the bandwidth allocated per user. VE was optimised to maximise acceptance ratio (the number of SFCRs that can be embedded divided by the total number of SFCRs received), and minimise link bottleneck and latency by Gamal et al.~\cite{Gamal2019, Gamal2019a}. Qu et al.~\cite{Qu2016} optimised VE to minimise latency. Mixed Integer Linear Programming, GA and Bee colony optimisation algorithms were used to optimise VE, and the Dijkstra algorithm was used to solve LE to minimise cost and latency by Khoshkholghi et al.~\cite{Khoshkholghi2020}. Kim et al. minimised power consumption by optimising VE and LE implicitly~\cite{Kim2016}. Rankothge et al. minimised the number of deployed servers, links, average link utilisation, and reconfiguration cost by optimising VE and LE explicitly~\cite{Rankothge2017}. VE was optimised, and LE was solved using k-shortest paths and first-fit techniques to maximise acceptance ratio and minimise CPU usage by Ruiz et al.~\cite{Ruiz2020}. Cao et al. optimised CC, VE, and LE sequentially to minimise link utilisation~\cite{Cao2017}. They took an implicit approach to LE. VE was optimised to maximise physical resource utilisation and minimise the number of hosts used by Tavakoli-Someh et al.~\cite{Tavakoli-Someh2019}. Fulber-Garcia et al. maximised inter-domain bandwidth availability, and minimised inter-domain latency, deployment cost and resource usage cost by optimising VE~\cite{Fulber-Garcia2024BreakingOrchestrators}. In another work, the authors also maximised the density of users using a multimedia service of a data centre and the geographical distance between two caches by optimising VE~\cite{Fulber-Garcia2023CustomizableMetaheuristics}.  Krishnamohan et al. optimised VE and solved LE using Dijkstra to maximise acceptance ratio and minimise latency, and used an SFC emulator named OpenRASE to evaluate the fitness of their solution~\cite{OpenRASE}. Unlike other works that use GAs, Krishnamohan et al. evaluated the fitness of individuals via online experimentation~\cite{10255468} on an emulator, which is claimed to offer better accuracy~\cite{Fahmy2023SimulatorsWSNs}.  To reduce the convergence time, they introduced a hybrid approach to optimise VE~\cite{Krishnamohan2025BeNNS:Embedding}. The hybrid approach used a surrogate model in tandem with OpenRASE to evaluate fitness. 

As can be seen from Table \ref{tab:ga_sfc_summary}, no study optimises all three sub-problems simultaneously using GA, underlining the research gap addressed by this paper.

\begin{table}
    \centering
    \begin{tabular}{ccccc}
         \hline
         \textbf{Study}&  \textbf{CC}&  \textbf{VE} &\textbf{LE}&  \textbf{Simultaneous}\\
         \hline
          
         \cite{Khoshkholghi2019}& X&  \checkmark &X&  N/A\\
         \cite{Carpio2017}&  X&  \checkmark &\checkmark (Implicit)&  \checkmark\\
         \cite{10.1007/978-3-031-33743-7_39}&  X&  \checkmark & X &  N/A \\
         \cite{Toumi2022}& X& \checkmark& \checkmark (Implicit)& \checkmark \\
         \cite{Gamal2019}&  X&  \checkmark &X&  N/A\\
         \cite{Gamal2019a}&  X&  \checkmark &X&  N/A\\
 \cite{Qu2016}& X& \checkmark &X& N/A\\
 \cite{Khoshkholghi2020}& X& \checkmark &\checkmark (Explicit)& X\\
 \cite{Kim2016}& X& \checkmark &\checkmark (Implicit)& X\\
         \cite{Rankothge2017}&  X&  \checkmark &\checkmark (Explicit)&  X\\
         \cite{Ruiz2020}&  X&  \checkmark &\checkmark (Explicit)&  \checkmark\\
         \cite{Cao2017}&  \checkmark&  \checkmark &\checkmark (Implicit)&  X\\
         \cite{Tavakoli-Someh2019}& X & \checkmark & No & N/A \\
         \cite{Fulber-Garcia2024BreakingOrchestrators}&X  & \checkmark & X  &N/A \\
         \cite{Fulber-Garcia2023CustomizableMetaheuristics}& X & \checkmark & X& N/A \\
 \cite{OpenRASE}& X& \checkmark& \checkmark (Explicit)& X\\
 \cite{Krishnamohan2025BeNNS:Embedding}& X& \checkmark& \checkmark (Explicit)& X\\
 \hline
 \\
    \end{tabular}
    \caption{A summary of other studies that optimise OSE using GAs.}
    \label{tab:ga_sfc_summary}
\end{table}

\subsection{GA and NN-based Algorithms}\label{ga_nn_rel}
Using typical machine learning and NN algorithms, such as gradient descent or backpropagation, to optimise NNs becomes complicated in GENESIS because: 
\begin{enumerate}
    \item Optimising the gradients of three DNNs simultaneously is complicated.
    \item Deriving a gradient for the fitness function is not straightforward since we use hybrid evolution (Section \ref{hybrid}).
\end{enumerate}
Furthermore, GAs are preferred instead because:
\begin{enumerate}
    \item GAs explore the search space efficiently, which helps avoid local optima~\cite{Ahmad2010PerformanceTraining}. 
    \item GAs enable concurrent search space exploration~\cite{Salomon1998EvolutionaryDifferences}.
\end{enumerate}

Using GAs to optimise the parameters and hyperparameters of NNs is a well-established~\cite{Whitley1990GeneticConnectivity}, common practice~\cite{Abdolrasol2021ArtificialReview}, with NeuroEvolution of Augmenting Topologies (NEAT)~\cite{stanley:ec02} being popular. This method has been used for time-series forecasting~\cite{ErzurumCicek2021OptimizingForecasting}, photonic device design~\cite{Ren2021Genetic-algorithm-basedDesign}, combinatorial optimisation~\cite{Shao2023Multi-ObjectiveProblems}, and optical neural network~\cite{Zhang2021EfficientAlgorithm}. Also, Gupta et al.~\cite{Gupta2019MemeticComputation} proposed the use of NNs for search-space dimensionality compression in GAs. Here, the gradients of the NN become the genetic encoding of the problem, and the output of the NN is used to produce a solution. 

We use the sine function as the activation function in GENESIS to increase the diversity of candidate solutions and enable more exploration (Section \ref{activation_func}). Using the sine activation function in NNs is not popular, but this approach has been used in implicit neural representation and short-term wind power forecasting~\cite{10.5555/3495724.3496350, Liu2021Short-termFunction}. 

To the best of our knowledge, this is the first work to use GAs and NNs to evolve a Gaussian distribution. While there exist probabilistic model-based optimisation algorithms that evolve a probabilistic distribution for an optimisation problem~\cite{Gupta2019MemeticComputation}, they do not evolve the parameters of the distribution; instead, they build the distribution of the parent population from the fittest individuals. 

\section{GENESIS Framework}\label{GENESIS}
We now introduce our GA-based approach to simultaneously optimising the OSE problem and its three sub-problems \footnote{GENESIS implementation can be found here: \url{https://github.com/Project-Kelvin/OpenRASE/tree/main/src/algorithms/hybrid}}. Formulating a genetic encoding for the OSE problem is challenging (Section \ref{gene_enc}). Therefore, we propose using the gradients of three NNs, which use the sine activation function to increase exploration (Section \ref{activation_func}), as the encoding to optimise the OSE problem (Section \ref{gene_enc}). Each of these NNs is contained within a solver that generates a candidate solution to one of the CC (Section \ref{cc_solver}), VE (Section \ref{ve_solver}), or LE (Section \ref{le_solver}) sub-problems. We optimise the gradients of the NNs simultaneously using the hybrid evolution approach (Section \ref{gen_evo}). 

\subsection{Genetic Encoding Challenges}\label{gene_enc}
To use GA, we devise a genetic encoding to represent candidate solutions. This is important because the encoding determines the effectiveness of the GA. However, this involves two major challenges:
\begin{enumerate*}
\item   developing a unified genetic encoding scheme to represent solutions to all three OSE sub-problems, and
\item search-space dimensionality compression
\end{enumerate*}.

The OSE problem consists of three optimisation sub-problems which have to be optimised \textit{simultaneously} instead of sequentially~\cite{GilHerrera2016}. In sequential optimisation, we first determine the order of VNFs in SFCs by optimising the CC problem, then decide which hosts the ordered VNFs should be embedded on by optimising VE, and finally, determine how the chosen hosts should be linked by optimising LE. During such sequential optimisation, the search space of the subsequent sub-problem is limited by the solution to the preceding sub-problem. For instance, a sub-optimal solution to VE may still produce an overall optimal solution~\cite{GilHerrera2016}; however, a sequential approach will limit itself to the optimal solutions to the VE sub-problem in the search space and not explore outside of it, risking an overall sub-optimal solution. 

In contrast, a simultaneous optimisation optimises all three sub-problems together, so it explores the entire search space. Meaning, an ideal encoding scheme should encode a candidate solution to all three optimisation problems. This encoding is further complicated by the fact that the three OSE optimisation sub-problems are of different types. The CC problem is an optimal ordering problem, VE is an optimal resource allocation problem, and LE is an optimal path finding problem. 

The two-dimensional binary encoding scheme used in the literature~\cite{OpenRASE, Krishnamohan2025BeNNS:Embedding, Cao2017} works well for resource allocation problems like VE, but adapting it to optimise ordering and path-finding problems is challenging. This binary encoding scheme is a function of the number of servers in the network topology and the number of VNFs in the SFCRs received. When the number of SFCRs and servers increases, so does the size of the encoding, increasing the dimensions of the search space. This can curtail the exploratory behaviour of GA, as it may not be possible to cover the entire search space with a reasonable population size and typical crossover and mutation strategies, affecting scalability. Therefore, an effective encoding scheme also has to compress the dimensions of the search space. 

\subsection{Genetic Encoding and Decoding}
\begin{figure}
    \centering
    \includegraphics[width=1\linewidth]{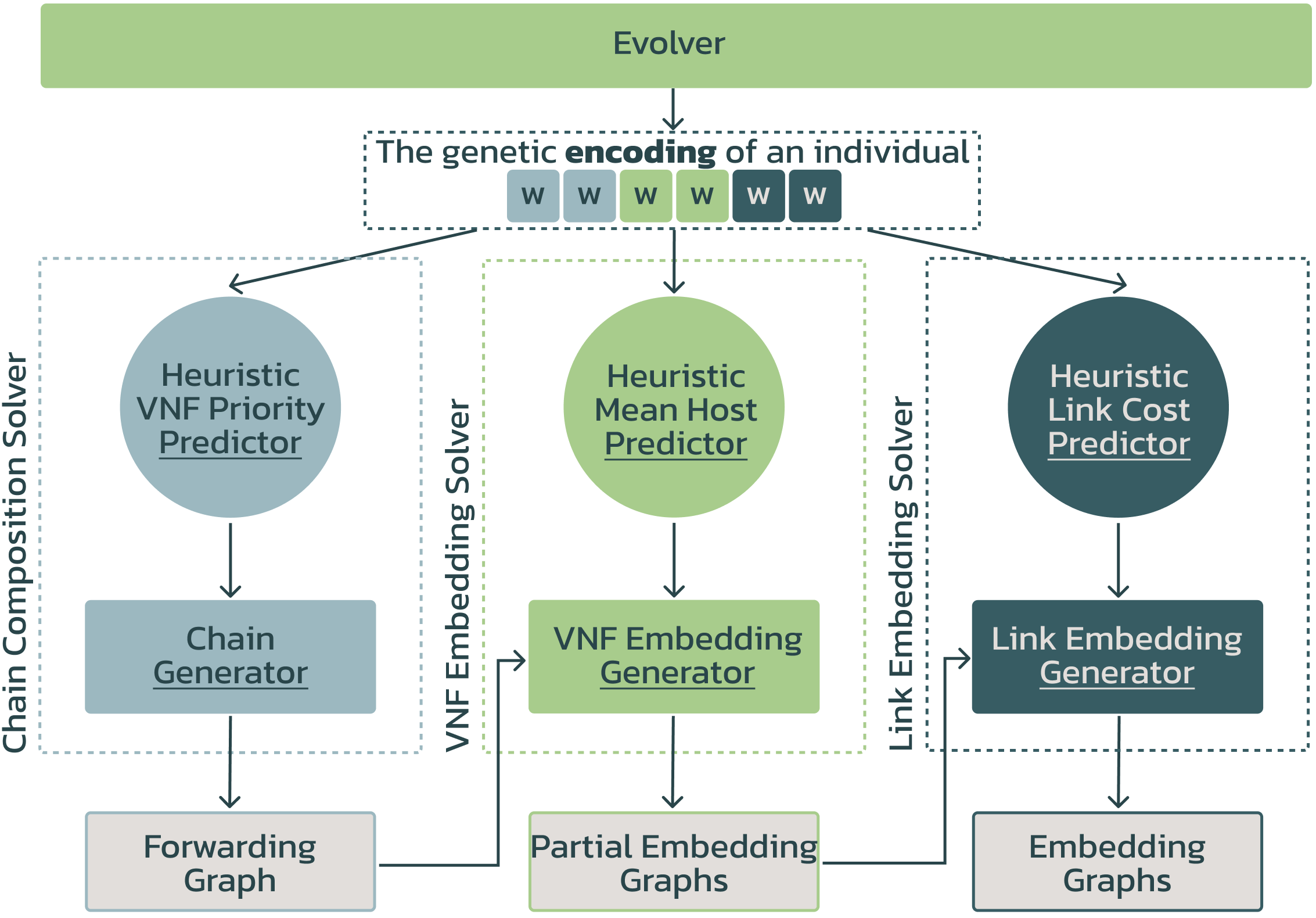}
    \caption{The evolver uses the gradients of three NNs as the genetic encoding of the GA to optimise the OSE sub-problems simultaneously. Three solvers use the output of the NNs to decode the encoding and generate a deployable candidate solution. The process of decoding is sequential; however, the gradients are optimised simultaneously.}
    \label{fig:solvers}
\end{figure}
We propose evolving three solvers, each containing a NN, to provide a unified encoding for all three sub-problems while compressing the dimensions of the search space. Fig.~\ref{fig:solvers} shows the end-to-end process of how an evolvable encoding is decoded to a deployable candidate solution
so that the OSE problem can be optimised. 

We now describe the process of decoding. GENESIS uses three solvers to decode the encoding and generate a candidate solution to the OSE problem, with a solver for each of the three sub-problems. A solver consists of a predictor and a generator. A predictor is a feed-forward, fully connected Deep Neural Network (DNN) that predicts a heuristic value, which is used by a generator to produce a candidate sub-solution. The solvers interact to produce a complete candidate solution. The CC solver produces the Forwarding Graphs (FGs), which contain the order of VNFs in an SFC. The VE solver takes the FGs and produces the Partial Embedding Graphs (PEGs), which contain the hosts in which the VNFs in the FGs should be deployed. The LE solver takes the PEGs and produces the Embedding Graphs (EGs), which contain the routing path between the VNF hosts in the PEGs. 
\begin{figure*}
    \centering
    \includegraphics[width=1\linewidth]{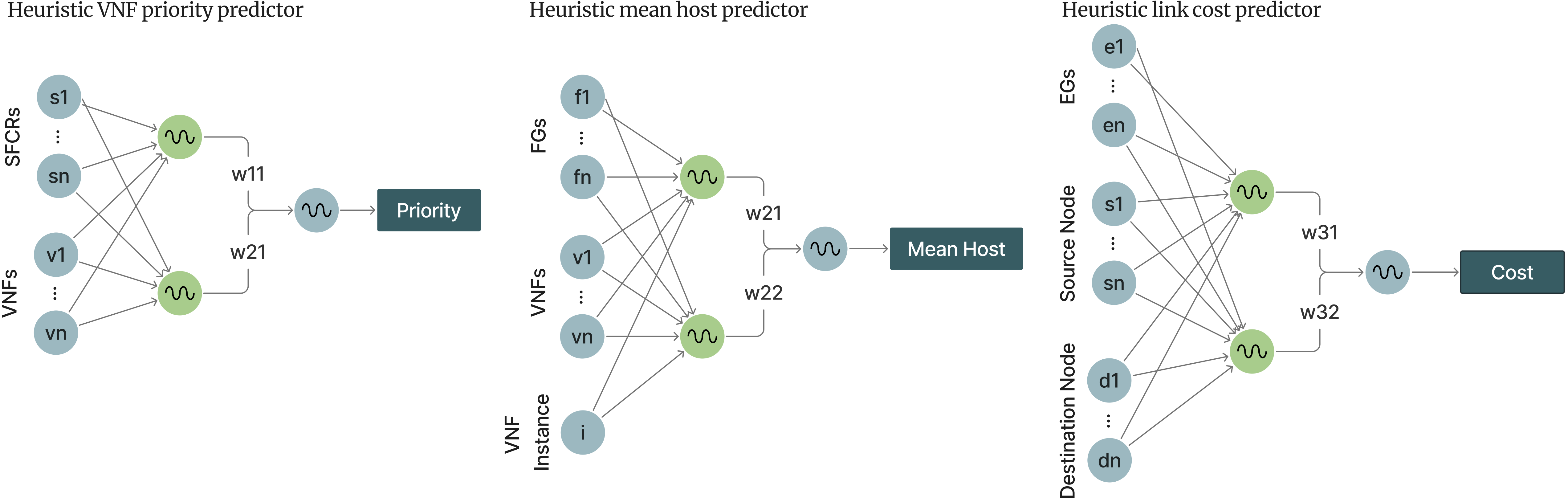}
    \caption{The predictors are fully connected, feed-forward DNNs with one hidden layer consisting of two neurons. The neurons use the sine activation function to increase exploration.}
    \label{fig:dnns}
\end{figure*}
All three DNNs consist of one hidden layer with two neurons, as shown in Fig. \ref{fig:dnns}. We generate the gradients of the input layer randomly and keep them constant throughout all generations of the evolution. We use GA to evolve the gradients of the hidden layer of all three DNNs simultaneously. Hence, we evolve $2\times3 = 6$ gradients simultaneously. Accordingly, a floating-point array with 6 elements becomes the genetic \textbf{encoding} for all three OSE sub-problems, as shown in Fig. \ref{fig:solvers}. It should be noted that even though the process of decoding is sequential, we evolve the gradients of the NNs simultaneously. Consequently, we optimise all three sub-problems simultaneously and take a sequential approach only to decode the encoding to generate deployable candidate solutions.  

This encoding also compresses the dimensions of the search space compared to the two-dimensional binary-encoding scheme (Section \ref{gene_enc}) and ensures that the dimensions of the search space do not grow as the number of SFCRs and the size of the topology increase, making this encoding scheme independent of the number of SFCRs and the network topology, ensuring scalability. For example, when evolving an optimal embedding scheme for 10 SFCRs with 2 VNFs each and a network topology with 10 hosts, the two-dimensional binary-encoding scheme produces an array of size $20 \times10$. In comparison, GENESIS produces an array of size $1\times6$. When we increase the number of SFCRs to 20, the size of the two-dimensional binary-encoding scheme increases to $40\times10$, whereas GENESIS's size remains at $1\times6$. It should be noted that the size of the input layers of GENESIS's NNs goes up with the increase in the number of SFCRs and the size of the topology, but this does not affect the encoding as the gradients of the input layers are randomly generated. 

\subsection{Activation Function of DNNs}\label{activation_func}
We use the sine function as the activation function in the DNNs instead of conventional activation functions, such as ReLU, to avoid dominant gradients inhibiting explorative behaviour. When using activation functions like ReLU, a few VNFs consistently appear first in the SFCs, and most VNFs are embedded on a few hosts. When using the sine function, the order of VNFs in SFCs becomes more diverse, and the VNFs are embedded across a wide range of hosts. 

\begin{figure}
    \centering
    \includegraphics[width=1\linewidth]{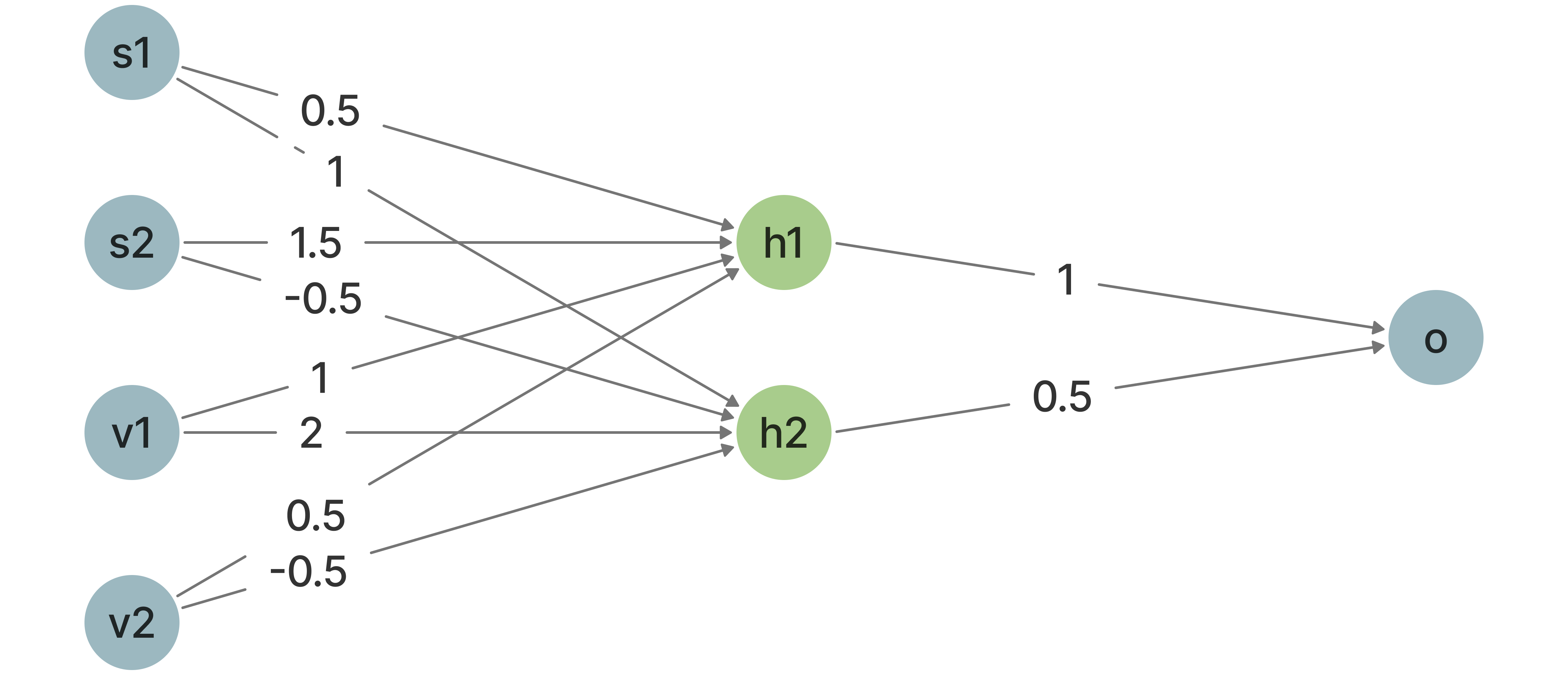}
    \caption{A DNN with gradients specified. V1's gradients are greater than V2's. }
    \label{fig:dom_grad}
\end{figure}

\paragraph{Example}
The dominant gradient problem can be exemplified using Fig. \ref{fig:dom_grad} and Table \ref{tab:dom_grad}. Let us assume that we need to find the optimal order of VNFs in two SFCs with two VNFs each. The output of the DNN is used to decide the order of VNFs in each SFC. $s1$ and $s2$ in the DNN shown in Fig. \ref{fig:dom_grad} are used to input one-hot encoded SFCRs. $v1$ and $v2$ are used to input VNFs. As shown in Fig. \ref{fig:dom_grad}, the gradients of $v1$ are greater than those of $v2$. When we use the ReLU activation function, as shown in Table \ref{tab:dom_grad}, VNF 1 produces the greater value for both SFCs, thanks to $v1$'s connections to the hidden layer having greater gradient values than $v2$. This means for both SFCs, VNF 1 is placed first. Similar behaviours were observed when experimenting with many SFCs: all SFCs almost had the same order of VNFs. This greatly reduces exploration in GA, risking getting stuck at local optima. To address this, we use the sine function as the activation function. The output of the sine function rises and falls as the input value increases, unlike ReLU, where the output has a linear relationship with the input. Consequently, as shown in Table \ref{tab:dom_grad}, when we use sine, for SFC 1, VNF 2 produces the greater value, whereas for SFC 2, VNF 1 produces the greater value. This mitigates the dominant gradient problem, enabling greater diversity in the population, which in turn encourages more exploration. We generate the initial gradients by randomly sampling from a uniform distribution between $-\pi$ and $\pi$, so that we get gradients along a full sine wave.

\begin{table}
    \centering
    \begin{tabular}{cccccc}
    \hline
         \textbf{s1}&  \textbf{s2}&  \textbf{v1}&  \textbf{v2}&  \textbf{Output with ReLU}&  \textbf{Output with Sine}\\ \hline
         1&  0&  1&  0&  3.000&  0.876\\
         1&  0&  0&  1&  1.250&  0.883\\
         0&  1&  1&  0&  3.250&  0.890\\
         0&  1&  0&  1&  2.000&  0.469\\         
         \hline
         \\
    \end{tabular}
    \caption{Outputs for ReLU and sine activation functions.}
    \label{tab:dom_grad}
\end{table}

\subsection{Chain Composition Solver}\label{cc_solver}
The CC solver optimises the CC problem. It consists of the Heuristic VNF Priority Predictor (HVPP) and the chain generator, as shown in Fig. \ref{fig:solvers}. 

\subsubsection{Heuristic VNF Priority Predictor}
This is a DNN that takes one-hot encoded SFCRs and VNFs as inputs and outputs a heuristic priority value, as shown in Fig. \ref{fig:dnns}. 
\subsubsection{Chain Generator}\label{chain_gen}
\begin{algorithm}
\small
    \caption{Chain Generator Algorithm using HVPP }\label{alg:vnf-cc}
    \begin{algorithmic}[1]
        \Require $SFCRs$: an array of SFCRs, $vnfs_i$: an array of VNFs in $SFCR_i$, $strict\_vnf\_order_i$: An array containing VNFs in the order they should appear in $SFCR_i$
        \Ensure $ordered\_vnfs\_in\_sfcs$: an array containing arrays of VNFs in the order they should appear in each SFC
        \State $ordered\_vnfs\_sfcs \gets \emptyset$
        \For {$SFCR_i \in SFCRs$}
            \State $ordered\_vnfs \gets \emptyset$
            \For{$vnf \in vnfs_i$}
                \State $priority \gets HVPP(vnf, SFCR\_i)$
                \State $vnf.priority \gets priority$
                \State $ordered\_vnfs.append(vnf)$
            \EndFor
            \State Sort $ordered\_vnfs$ by descending order using $vnf.priority$
    
            \State $last\_index \gets -1$
            \For{$strict\_vnf \in strict\_vnf\_order_i$}
                \State $index \gets ordered\_vnfs.index(strict\_vnf)$
                \If{$index < last\_index$}
                    \State $ordered\_vnfs.remove(strict\_vnf)$
                    \State $index \gets last\_index$
                    \State $ordered\_vnfs.insertAt(index, strict\_vnf)$
                \EndIf
                \State $last\_index \gets index$
            \EndFor
            \State $ordered\_vnfs\_sfcs.append(ordered\_vnfs)$
        \EndFor
        \State \Return{$ordered\_vnfs\_sfcs$}
    \end{algorithmic}
\end{algorithm}
The chain generator arranges the VNFs in each SFCR in the descending order of their priority values according to Algorithm \ref{alg:vnf-cc}. If an SFCR specifies a strict order among certain VNFs, then the chain generator ensures that VNFs appear according to the strict order. The output of the chain generator is the FGs, which specify the order of VNFs in SFCs. They are used as an input to the VE solver. 

\subsection{VNF Embedding Solver}\label{ve_solver}
\begin{figure*}[t]
    \centering
    \includegraphics[width=0.8\linewidth]{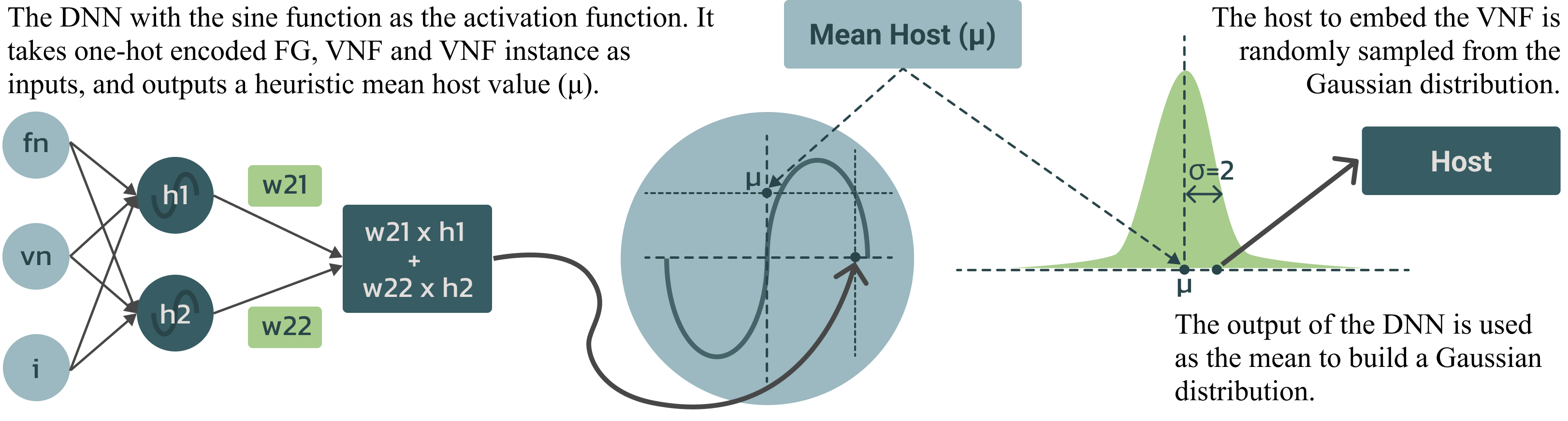}
    \caption{A diagram depicting the process of translating genetic encoding ($w1$ and $w2$) to decide the host to embed a VNF on.}
    \label{fig:gauss_dis}
\end{figure*}
The VE solver optimises the VE problem of the OSE problem. This solver consists of the Heuristic Mean Host Predictor (HMHP) and the VNF embedding generator, as shown in Fig. \ref{fig:solvers}.

\subsubsection{Heuristic Mean Host Predictor}\label{hmhp}
This is a DNN that accepts one-hot encoded FGs and VNFs, and the VNF instance number as inputs and outputs a heuristic mean host value for each VNF of an FG, as shown in Fig. \ref{fig:dnns}. The VNF instance number is used to uniquely identify each instance of a VNF in an FG. Multiple instances of a VNF can be found in SFCs that contain VNFs that split the chain, such as a load balancer. The amplitude of the sine activation function of the output neuron is set to the number of hosts available, as shown in Equation \ref{host_sine_func}.
\begin{equation}\label{host_sine_func}
    mean\_host = no.\ of\ hosts \times sin(w_{21}\cdot h_1 + w_{22} \cdot h_2)
\end{equation}

\subsubsection{VNF Embedding Generator}\label{vnf_em_gen}
\begin{algorithm}
\small
    \caption{VNF Embedding Generator Algorithm using HMHP}\label{alg:vnf-embedding}
    \begin{algorithmic}[1]
        \Require $FGs$: an array of FGs,
        $vnfs_i=\{(vnf, instance\_id)_n\}$: an array of tuples containing the VNF type and its instance ID in $FG_i$, $n$: number of hosts in the topology
        \Ensure $vnf\_hosts=\{(FG_i, vnf, instance\_id, host)_n\}$: an array of tuples containing the host in which the specific VNF should be deployed
        \State $vnf\_hosts \gets \emptyset$
        \For $FG_i \in FGs$
            \For{$vnf, instance\_id \in vnfs_i$}
                \State $mean\_host \gets HMHP(vnf, instance\_id,$ \linebreak $FG_i)$
                \If{$mean\_host > 0$}
                    \State $random\_host \gets$ $\mathcal{N}$$(mean\_host, 2^2)$
                    \State $host = floor(random\_host)$
                    \State $host = host\mod n$
                    \If{$host <  0$}
                        \State $host = n + host$
                    \EndIf
                    \State $vnf\_hosts.append(FG_i, vnf, \linebreak instance\_id, host)$
                \Else
                    \State $vnf\_hosts.append(FG_i, vnf,
                    \linebreak instance\_id, None)$ 
                \EndIf
            \EndFor
        \EndFor
        \State \Return{$vnf\_hosts$}
    \end{algorithmic}
\end{algorithm}

The VNF embedding generator takes the FGs produced by the CC solver (Section \ref{cc_solver}) and adds the host in which each VNF should be deployed based on the heuristic mean host value predicted by HMHP, as described by Algorithm \ref{alg:vnf-embedding}, to produce PEGs, which are used by the link embedding solver to embed links. The generator decides the host of a VNF using the heuristic mean host. Since we set the amplitude of the sine activation function of the output neuron of the DNN to the number of hosts in the topology (Section \ref{hmhp}), and reject VNFs with a heuristic mean host less than 0, the heuristic mean host is a value between 0 and the number of hosts. To encourage more exploration, the generator builds a Gaussian distribution with the mean ($\mu$) set to the heuristic mean host and the standard deviation ($\sigma$) set to $2$, as shown in Fig. \ref{fig:gauss_dis}. Then, the generator samples a random value from the distribution and finds its floor to generate the index of the host for the concerned VNF. The generator uses the modulo to find the index if the random value exceeds the limits. $\sigma$ is a tunable hyperparameter. Setting it to a higher value increases the spread of the distribution, producing a more explorative behaviour. A lower value narrows the distribution, resulting in a more exploitative behaviour. 

The VE solver, using a DNN, reduces the VNF embedding problem to a numerical value, which is used to produce a Gaussian distribution. In effect, the GA evolves an optimal Gaussian distribution, from which the host to embed a VNF on is randomly sampled.

\subsection{Link Embedding Solver}\label{le_solver}
The link embedding solver optimises the LE problem of the OSE problem. It consists of the Heuristic Link Cost Predictor (HLCP) and the link embedding generator.

\subsubsection{Heuristic Link Cost Predictor}
This is a DNN that takes the PEGs, the source node and the destination node as inputs and outputs the heuristic cost between the two nodes for a given PEG, as shown in Fig. \ref{fig:dnns}. The nodes consist of the hosts and network switches in the topology. 

\subsubsection{Link Embedding Generator}
\begin{algorithm}
\small
    \caption{Link Embedding Generator Algorithm using HLCP}\label{alg:link-embedding}
    \begin{algorithmic}[1]
        \Require $EGs$: an array of EGs, $src$: the source node, $dst$: the destination node
        \Ensure $optimal\_path\_sfcs$: an array of arrays containing the nodes in the optimal path between source and destination for each EG

        \State $optimal\_path\_sfcs \gets \emptyset$
        \For {$EG_i \in EGs$}
            \State $open\_set \gets [src]$
            \State $closed\_set \gets \emptyset$
            \State $index \gets 0$
    
            \While{$length(open\_set) > 0$}
                \State $curr\_node \gets getLeastTotalCostNode(open\_set)$
                \If{$curr\_node == dst$}
                    \State $path \gets tracePath(curr\_node)$\Comment{using $parent$ attribute}
                    \State $optimal\_path\_sfcs.append(optimal\_path)$

                    \State \textbf{break}
                \EndIf
    
                \If{$index == 0$ or not $isHost(curr\_node)$}
                    \For{$nb \in getNeighbours(curr\_node)$}
                        \State $nb.costToDst \gets HLCP(EG_i,nb,dst) $
                        \State $costFromCurrNode \gets HLCP(EG_i, \linebreak curr\_node, nb)$
                        \State $nb.costFromSrc \gets nb.costFromSrc \linebreak + costFromCurrNode$
                        \State $nb.totalCost \gets nb.costToDst \linebreak + nb.costFromSrc$
                        \State $nb.parent \gets curr\_node$
    
                        \If{$nb \not \in closed\_set$}
                            \State $open\_set.append(nb)$
                        \ElsIf{$nb.totalCost < closed\_set.get(\linebreak nb).totalCost$}
                            \State $open\_set.append(nb)$
                        \EndIf
                    \EndFor
                \EndIf
                \State $index ++$
                \State $closed\_set.append(curr\_node)$
            \EndWhile
    \EndFor
    \State \Return{$optimal\_path\_sfcs$}
    \end{algorithmic}
\end{algorithm}

The link embedding generator adapts the A* algorithm~\cite{Hart1968APaths} to find an optimal path between the hosts of two successive VNFs. The predictor is used to find the heuristic cost between two nodes (hosts and switches) in the network, as explained in Algorithm \ref{alg:link-embedding}. The generated least-cost links are embedded on the PEGs (Section \ref{vnf_em_gen}), producing EGs. The EGs are used by the OpenRASE SFC emulator to embed and evaluate the SFCs on the network.

By using the A* algorithm to find an optimal path between two hosts \textit{via switches}, we explicitly optimise LE. Moreover, unlike other explicit approaches that connect hosts via the shortest path, the heuristic cost produced by the predictor enables hosts to be connected via different paths for different SFCs, mitigating traffic congestion and improving resource utilisation.

\subsection{Genetic Evolution}\label{gen_evo}
We evolve the gradients of the hidden layer of the DNNs using GA. An individual in this case is a floating-point array with 6 elements, as shown in Fig. \ref{fig:solvers}. We first generate individuals randomly from a uniform distribution between $-\pi$ and $\pi$ to obtain outputs along a complete sine wave. We then use hybrid evolution (Section \ref{hybrid}) to evolve these gradients. The GA we use to evolve these gradients takes the following steps:
\begin{enumerate*}
    \item \textit{Initial population generation}--Individuals are randomly generated using a uniform distribution with each gene taking a value between  $-\pi$ and $\pi$. 
    \item \textit{Evaluation}--Each individual in the population is evaluated first by decoding using the solvers to produce the EGs. Then the fitness of the EGs is evaluated using the hybrid evolution approach (Section \ref{hybrid}).
    \item \textit{Crossover}--Two random individuals are subjected to blend crossover~\cite{ESHELMAN1993187} to produce offspring. Blend crossover was chosen because a gene in an individual is a floating-point number. The alpha value of the blend crossover was set to 0.5, which is the typical value.
    \item \textit{Mutation}--All individuals are subjected to mutation. Mutation happens at the gene level using a Gaussian distribution with a mean of 0 and a standard deviation of $\pi$.
    \item \textit{Selection}--We select the next generation of individuals from the parent and offspring population using a selection strategy that is use-case dependent. 
\end{enumerate*}

\subsubsection{Hybrid Evolution}\label{hybrid}
Hybrid evolution combines offline fitness evaluation, which uses a surrogate model to approximate fitness, with online evaluation, which involves experimentally evaluating fitness on the SFC emulator OpenRASE~\cite{OpenRASE} to provide accurate fitness evaluation at an acceptable speed~\cite{Krishnamohan2025BeNNS:Embedding}. It evolves using a surrogate model until the fitness meets a threshold, after which it evaluates the fitness on OpenRASE. If the fitness measured on OpenRASE does not meet the threshold, then evolution resumes using the surrogate model. 

\section{Evaluation}\label{eval}
\subsection{Experiment Setup}
We carried out the experiments on a virtual machine running Ubuntu 20.04.6 on a QEMU hypervisor with 64 cores of Intel Xeon Gold 6240R CPUs having a clock speed of 2.4 GHz, and 64 GB of RAM.

\subsection{Experiment Use Case}
To evaluate GENESIS, we chose a data centre environment since it is a popular use case in the literature for the OSE problem~\cite{Xie2021, Jang2017, Xu2021, Wang2020a}. A 4-ary fat-tree topology was used for the substrate network, as it is a typical data centre topology~\cite{AlqahtaniRethinkingCenters}. 

\subsection{Optimisation Objectives}
The optimisation problem is a Pareto optimisation problem where the \textit{acceptance ratio} has to be maximised, and the \textit{average traffic latency} has to be minimised. Acceptance ratio is the number of SFCRs that can be embedded divided by the total number of SFCRs received. Traffic latency is the amount of time taken by a request to traverse an SFC. We average the traffic latencies of all SFCs embedded to find the average traffic latency. These are conflicting objectives. When we accept to embed more SFCRs, the competition for network resources goes up, increasing the traffic latency. We used the NSGA-II algorithm~\cite{Deb2002ANSGA-II} for selection (Section \ref{gen_evo}). We can mathematically compute the acceptance ratio, but we have to find the traffic latency using the hybrid approach (Section \ref{hybrid}), as traffic latency has to be either measured or estimated using a model.

\subsection{Network Configurations}
We set the memory available to a host in the topology to 5 GB across all experiments, as memory was shown not to have an impact on traffic latency~\cite{Krishnamohan2025BeNNS:Embedding}. The CPUs available to hosts were set to 0.5, 1, and 2, and the bandwidth available to links was set to 5 MB and 10 MB across all experiments. We scaled down the resources available to emulate a miniaturised data centre, as replicating real resource availability was impractical in the test machine. 

\subsection{SFCRs}
We developed four unique SFCRs based on common VNFs used in data centres~\cite{Yang2016Energy-awareCenters, Herker2015Data-centerRequirements, Zu2021FairCenter}:
\begin{enumerate*}
    \item Load Balancer $\rightarrow $ Web Application Firewall
    \item HTTP Accelerator $\rightarrow$ Load Balancer $\rightarrow$ Web Application Firewall
    \item HTTP Accelerator $\rightarrow$ Traffic Monitor $\rightarrow$ Load Balancer $\rightarrow$ Web Application Firewall
    \item Load Balancer $\rightarrow$ Traffic Monitor $\rightarrow$ Web Application Firewall 
\end{enumerate*}. 
We made 8 and 12 copies of each of these SFCRs to adjust the number of SFCRs to be embedded during experiments. 

\subsection{Traffic Patterns}
The traffic pattern (Traffic A) used for the experiments was based on the traffic trace we obtained from a campus data centre network~\cite{Benson2010NetworkWild}. To create more traffic patterns, we scaled the amount of traffic by 2 (Traffic 2x) and phase shifted the traffic pattern by 50\% (Traffic B).  

\subsection{Comparisons}
To demonstrate the effectiveness of GENESIS, we compared three other heuristic approaches: the Binary-Encoded GA (BEGA)~\cite{Krishnamohan2025BeNNS:Embedding}, the GA-based Heuristic Algorithm (GAHA)~\cite{Khoshkholghi2019}, and the Greedy Dijkstra Algorithm (GDA)\footnote{\url{https://sourceforge.net/p/alevin/svn/HEAD/tree/trunk/src/vnreal/algorithms/samples/SimpleDijkstraAlgorithm.java}}. We chose BEGA and GAHA because they both aim to minimise the traffic latency, and chose GDA because it is a greedy algorithm that serves as a benchmark to compare the performance of GAs against. We did not compare GENESIS to the GA proposed by Cao et al.~\cite{Cao2017}, which optimises all three OSE sub-problems, because its fitness function had to be modified to accommodate the optimisation objectives of this study. Since the fitness function is a fundamental part of a GA, modifying it significantly alters the performance of the algorithm, precluding a fair comparison. 

Additionally, BEGA, GAHA and GDA optimised only the VE sub-problem of OSE. This allowed us to evaluate the performance advantage GENESIS gains by optimising all three OSE sub-problems. GAHA uses offline fitness evaluation, so it also allowed us to compare an offline approach to the hybrid approach of GENESIS and BEGA. We used the benchmarking models available in OpenRASE~\cite{OpenRASE} for estimating CPU utilisation of VNFs in GAHA because its offline approach lacks an inherent estimation mechanism.

\subsection{Experiment Configurations}
\begin{table}
    \centering
    \begin{tabular}{p{0.30\linewidth}p{0.10\linewidth}p{0.13\linewidth}p{0.13\linewidth}p{0.10\linewidth}}
        \hline
        \textbf{Parameter}& \textbf{GENESIS}& \textbf{BEGA 100}& \textbf{BEGA 2000}& \textbf{GAHA} \\
        \hline
        Min. Acceptance Ratio&1&1&1&1\\
        Max. Traffic Latency&100 ms&100 ms&100 ms&500 ms\\
        Population Size&100&100&2000&100\\
        Max. Generations&500&500&500&500\\
         \hline
         \\
    \end{tabular}
    \caption{The experimental parameters used to evaluate algorithms.}
    \label{tab:thresholds}
\end{table}

The convergence threshold (Section~\ref{hybrid}) for GENESIS and BEGA was set to a \textit{minimum acceptance ratio of 1} and a \textit{maximum average traffic latency of 100 ms}, as shown in Table \ref{tab:thresholds}. Since we found GAHA's offline approximation to not be accurate enough, with the difference between the average traffic latency measured online and approximated offline to be 519.28 ms (Section \ref{overall_res}), we set the maximum traffic latency for GAHA to 500 ms. We set the maximum allowed generations for all three algorithms to 500, so if the algorithm did not converge within 500 generations, the evolution was halted. We set the number of individuals in a population to 100. However, to determine if BEGA would perform better with more individuals in a population, as its binary encoding produces a search space with many dimensions, requiring more exploration, we ran another set of experiments with the population size set to 2000. We did not do the same with GAHA, as it was significantly slower compared to the other algorithms, even with only 100 individuals in a population (Section~\ref{results}). 

\subsection{Experiment Design}
We designed 48 experiments with varying traffic patterns, numbers of SFCRs, and topological configurations to evaluate GENESIS's performance and scalability. We ran these 48 experiments in two stages, with 24 experiments in each stage: first, with 32 SFCRs and second, with 48 SFCRs. In the first stage, we evaluated all GA-based algorithms and GDA. For the second stage, we only selected the GAs that converged in at least one experiment in the first stage, along with GDA. Each experiment was run only once to keep the time taken to evaluate one algorithm down to a practical level, as some algorithms, such as GAHA, took prohibitively long (709.36 minutes). 

\section{Results}\label{results}
\begin{figure*}[t]
    \centering
    \includegraphics[width=1\linewidth]{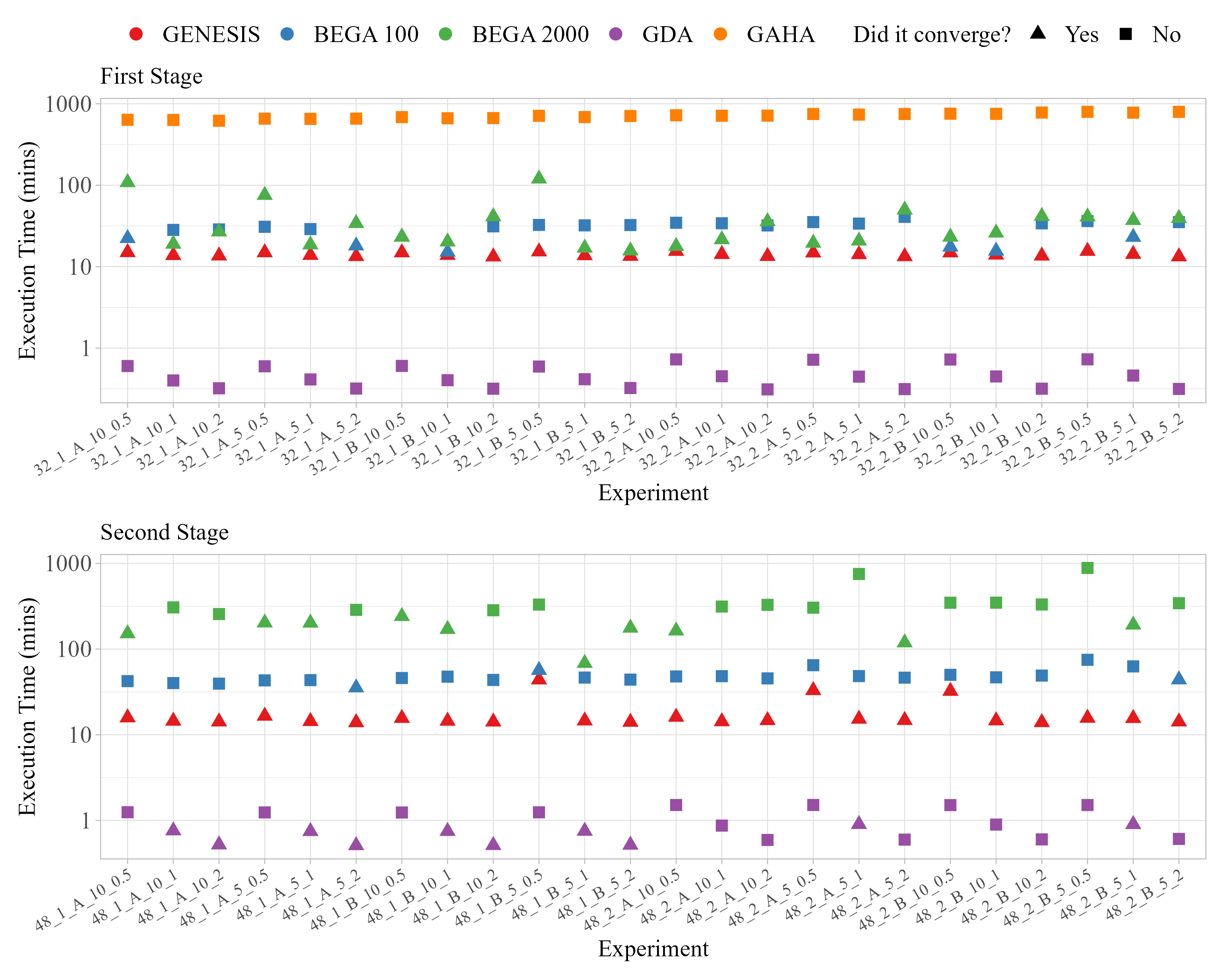}
    \caption{The total time taken for evolution in the first stage. The shape of the points is determined by whether or not the evolution converged on a solution that met the threshold. Each experiment is named after its configuration, punctuated by an underscore, in the following order: number of SFCRs, traffic scale, traffic pattern, bandwidth and number of CPUs.}
    \label{fig:first_evo_time}
\end{figure*}
\subsection{Convergence and Execution in First Stage}

Fig. \ref{fig:exec_time} shows the execution and convergence times in the first stage. GENESIS and BEGA with 2000 individuals (BEGA 2000) converged on a solution that met the threshold in all 24 experiments. BEGA with 100 individuals (BEGA 100) converged in 7 experiments, while GAHA and GDA failed to converge in any experiment. GDA was the fastest to finish across all experiments. Among GAs, GENESIS was the fastest, as shown in Fig. \ref{fig:first_evo_time}. BEGA 100 was faster than BEGA 2000 in all experiments where it converged. GAHA took an average of 709.36 minutes and was the slowest of all algorithms.

\subsection{Convergence and Execution in Second Stage}
Fig. \ref{fig:exec_time} shows the execution and convergence times in the second stage. For the second stage, only GENESIS, GDA, BEGA 100 and BEGA 2000 were used, as GAHA failed to converge in all experiments in the first stage and was the slowest. GENESIS converged in all 24 experiments, while BEGA 2000 converged in only 10 experiments. BEGA 100 converged in only 3 experiments. GDA converged in 10 experiments. GENESIS was the fastest across all experiments among GAs, while GDA was the fastest overall. 

\subsection{Overall Performance}\label{overall_res}
\begin{figure}
    \centering
    \includegraphics[width=1\linewidth]{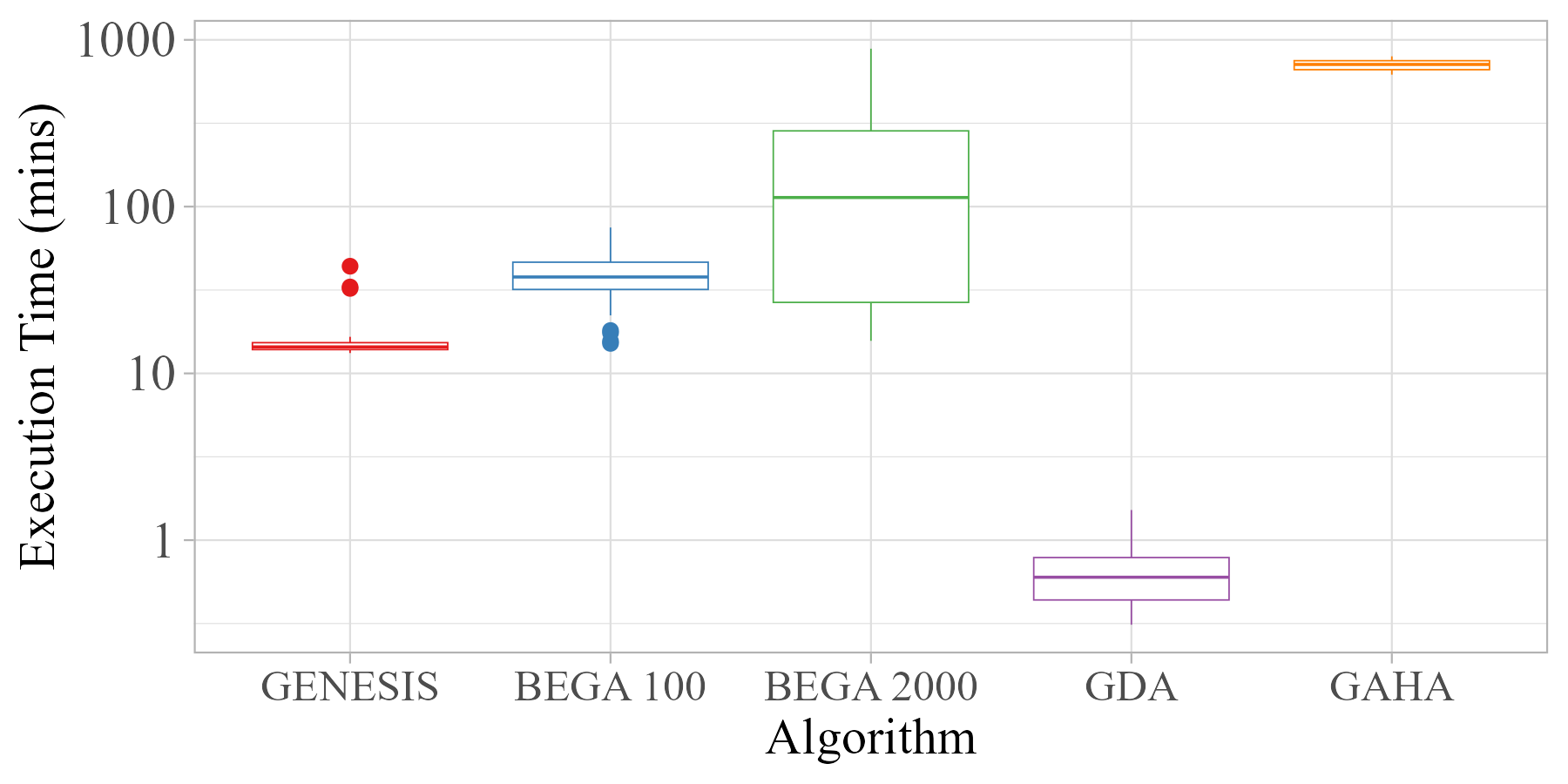}
    \caption{The average execution time of algorithms across all experiments.}
    \label{fig:exec_time}
\end{figure}

Across both stages, GENESIS converged in all 48 (100\%) experiments, while BEGA 2000 converged in 34 (71\%) experiments\footnote{Experiment data is available here: \url{https://github.com/Project-Kelvin/open-research-data/tree/main/genesis}}. BEGA 100 and GDA converged in 10 (21\%) experiments, while GAHA converged in none. Fig. \ref{fig:exec_time} shows the distribution of the execution times of the evaluated algorithms. A GENESIS experiment took an average of 15.84 (13.25-43.91) minutes, while BEGA 2000 took 166.64 (15.68-885.13) minutes, BEGA 100 took 38.62 (15.13-74.94) minutes, and GDA took 41.66 (18.6-91.2) s. GAHA took 709.36 (618.59-796.51) minutes. GENESIS was 2.4 times faster than BEGA 100, which was the second-fastest GA-based approach. GENESIS converged in 2.25 (1-9) generations on average, while BEGA 100 took 428.75 (23-500) generations and BEGA 2000 took 231.73 (7-500) generations. The mean difference between the traffic latency computed offline by GAHA and measured on OpenRASE was 519.28 (254.07-1021) ms. 
\section{Discussion}
The 48 experiments show that GENESIS was able to 
\begin{enumerate*}
    \item produce an optimal solution across different configurations, making it the most scalable algorithm overall, 
    \item produce an optimal solution faster than other GA-based approaches. 
\end{enumerate*}
This is because GENESIS optimises all three OSE sub-problems, unlike the other algorithms that optimise only the VE and LE sub-problems. This provides GENESIS more degrees of freedom, as it can optimise both the order of VNFs in SFCs and the physical paths between them, in addition to optimising where the VNFs are embedded. The use of an evolved Gaussian distribution to make VNF embedding decisions made GENESIS more effective than BEGA, which uses a two-dimensional binary array to make embedding decisions. Furthermore, using the A* algorithm with an evolved heuristic to explicitly optimise LE enabled GENESIS to outperform BEGA and GAHA, which use the Dijkstra algorithm to find the shortest path between hosts. 

Compressing the dimensions of the search space and enabling more exploration by using the sine activation in the DNNs, along with the use of a Gaussian distribution for VNF embedding, allowed GENESIS to evolve with just 100 individuals in a population, which greatly reduced the time taken to converge. BEGA 100, which also used only 100 individuals, was faster compared to BEGA 2000, but did not converge for most experiments because the binary encoding used in this algorithm produces a search space with many dimensions, requiring more extensive exploration. The comparatively better explorative quality of GENESIS enabled it to converge on an optimal solution in an average of 2.25 generations (Section~\ref{results}), whereas the other GAs took several hundred generations. 

GDA, being a greedy algorithm that involves no evaluation, was significantly faster than the other algorithms across all experiments. However, it did not converge in any experiment in the first stage and converged in 10 experiments in the second stage. GDA performed better in the second stage despite the number of SFCRs being higher. This was because GDA is a greedy algorithm that embeds SFCRs in the order they arrive. The order of SFCRs differed between the stages, as the first stage used 8 copies of each SFCR, while the second stage used 12 copies of each SFCR (Section~\ref{eval}). It is important to note that, unlike GENESIS and BEGA, GDA does not include a verification step, so the performance of the solution can only be known after embedding it on the \textit{production} environment. GAHA's offline evaluation of traffic latency was shown to be inaccurate, as there was a high discrepancy between offline and online evaluations, impacting GAHA's performance.

\section{Future Work}\label{future}
This paper focused on devising an encoding scheme to represent solutions to all three OSE sub-problems and tuning the hyperparameters of GENESIS, such as the mutation rate, the standard deviation of the Gaussian distribution, the architecture of the NNs, and the size of the population, for generalised scenarios. Evolving the hyperparameters to improve its performance further in specific scenarios, such as 48\_1\_B\_5\_0.5, 48\_2\_A\_5\_0.5, and 48\_2\_B\_10\_0.5, which took longer than other scenarios, will be an interesting future step. We plan to explore the use of a meta-evolver to evolve the hyperparameters. Additionally, the evaluations were carried out in static network environments; however, a real network environment is dynamic, where the SFCRs continue to arrive and the network topology changes. GENESIS's encoding scheme is agnostic of the number of SFCRs and the network topology, and hence, we believe GENESIS can be adapted to work in a dynamic environment. Evaluating GENESIS in a dynamic network environment is another work planned for the future. 

\section{Conclusion}
This paper presents a solution to the OSE problem by evolving three NNs using GA, called GENESIS, to optimise all three sub-problems simultaneously using a hybrid evolution approach. The NNs use the sine activation function to increase exploration. We use the output of one of the NNs to generate a Gaussian distribution to optimise the VE sub-problem, while using the output of another NN in the A* algorithm to optimise the LE sub-problem. Experimental comparisons of GENESIS across 48 scenarios with 2 state-of-the-art GAs and a greedy algorithm from the literature show that GENESIS is more effective and faster, producing an optimal solution in all 48 scenarios within a comparatively shorter time, while the other GAs failed to produce an optimal solution in all scenarios, despite taking longer. The greedy algorithm is faster than GAs, but it produced an optimal solution in only 10 scenarios. 

\section*{Acknowledgement}
This work was supported by EPSRC EP/X5257161/1. For the purpose of open access, the authors have applied a Creative Commons Attribution (CC BY) licence to any Author Accepted Manuscript version arising from this submission.

\printbibliography

\end{document}